\documentclass[journal]{IEEEtran}
\usepackage{graphicx}
\usepackage{mathrsfs}
\usepackage{color}
\usepackage{enumerate}
\usepackage{url} 
\usepackage{times}
\usepackage{amsmath}
\usepackage{booktabs}
\usepackage{diagbox}
\usepackage{times}
\usepackage{epsfig}
\usepackage{subfigure}
\usepackage{amssymb}
\usepackage{array}
\usepackage{multirow}
\usepackage{helvet}
\usepackage{courier}
\usepackage{enumitem}
\usepackage[ruled,linesnumbered]{algorithm2e}
\usepackage[pagebackref=true,breaklinks=true,citecolor=blue,linkcolor=blue,letterpaper=true,colorlinks,bookmarks=false]{hyperref}

\begin{document}
\title{ORDNet: Capturing Omni-Range Dependencies for Scene Parsing}
\author{Shaofei~Huang, Si~Liu, Tianrui~Hui, Jizhong~Han, Bo~Li, Jiashi~Feng and Shuicheng~Yan,~\IEEEmembership{Fellow,~IEEE}
\thanks{This research is supported in part by the National Key Research and Development Program of China under Grant 2020AAA0130200, National Natural Science Foundation of China under Grant 61876177, Beijing Natural Science Foundation under Grant 4202034. \textit{(Corresponding author: Si Liu)}}
\thanks{Shaofei Huang, Tianrui Hui, and Jizhong Han are with Institute of Information Engineering, Chinese Academy of Sciences, E-mail: \{huangshaofei, huitianrui, hanjizhong\}@iie.ac.cn.}
\thanks{Si Liu and Bo Li are with Institute of Artificial Intelligence, Beihang University, E-mail: \{liusi, boli\}@buaa.edu.cn.}
\thanks{Jiashi Feng is with Department of Electrical and Computer Engineering, National University of Singapore. E-mail: elefjia@nus.edu.sg.}
\thanks{Shuicheng Yan is with YITU Technology, Email: shuicheng.yan@yitu-inc.com.}}

\maketitle

\markboth{IEEE Transactions on Image Processing}%
{Shell \MakeLowercase{\textit{et al.}}: Bare Demo of IEEEtran.cls for IEEE Journals}


\begin{abstract}
    Learning to capture dependencies between spatial positions is essential to many visual tasks, especially the dense labeling problems like scene parsing. 
    Existing methods can effectively capture long-range dependencies with self-attention mechanism while short ones by local convolution. 
    However, there is still much gap between long-range and short-range dependencies, which largely reduces the models' flexibility in application to diverse spatial scales and relationships in complicated natural scene images.
    To fill such a gap, we develop a Middle-Range (MR) branch to capture middle-range dependencies by restricting self-attention into local patches.
    Also, we observe that the spatial regions which have large correlations with others 
    can be emphasized to exploit long-range dependencies more accurately, and thus propose a Reweighed Long-Range (RLR) branch. 
    Based on the proposed MR and RLR branches, we build an Omni-Range Dependencies Network (ORDNet) which can effectively capture short-, middle- and long-range dependencies. Our ORDNet is able to extract more comprehensive context information and well adapt to complex spatial variance in scene images. 
    Extensive experiments show that our proposed ORDNet outperforms previous state-of-the-art methods on three scene parsing benchmarks including PASCAL Context, COCO Stuff and ADE20K, demonstrating the superiority of capturing omni-range dependencies in deep models for scene parsing task.
\end{abstract}

\begin{IEEEkeywords}
Scene Parsing, Omni-Range Dependencies, Self-Attention
\end{IEEEkeywords}

\IEEEpeerreviewmaketitle

\section{Introduction}
\label{sec:introduction}

\IEEEPARstart{S}{cene} parsing~\cite{zhao2017pyramid}\cite{zhou2017scene}\cite{hung2017scene}\cite{zhang2017scale} aims to divide the entire scene into different segments and predict the semantic category for each of them. 
It is a fundamental task in computer vision and image processing, challenged by the complexity of natural scenes that usually contain multiple elements of various categories, including discrete objects (e.g., person, cat) and stuff (e.g., sky, river, grass). 
The elements within a scene may be spatially dependent upon each other. 
For example, a ship usually appears on the sea rather than on the road. 
Such dependencies of spatial positions can be exploited to boost the prediction accuracy further.

Mainstream scene parsing models built on Fully Convolutional Networks~\cite{Long_2015_CVPR} incorporate carefully designed modules to exploit spatial context information.
For example, Deeplabv2~\cite{chen2017deeplab} uses an Atrous Spatial Pyramid Pooling (ASPP) module to sample feature maps in parallel with different atrous rates to enlarge the receptive field of filters;
PSPNet~\cite{zhao2017pyramid} performs pooling operations at multiple grid scales for the same goal.
With a larger receptive field, these networks are able to extract broader scales of spatial context information.
However, these methods model the dependencies between positions in an implicit way.

\begin{figure}[t]
  \centering 
  \includegraphics[width=1.00\linewidth]{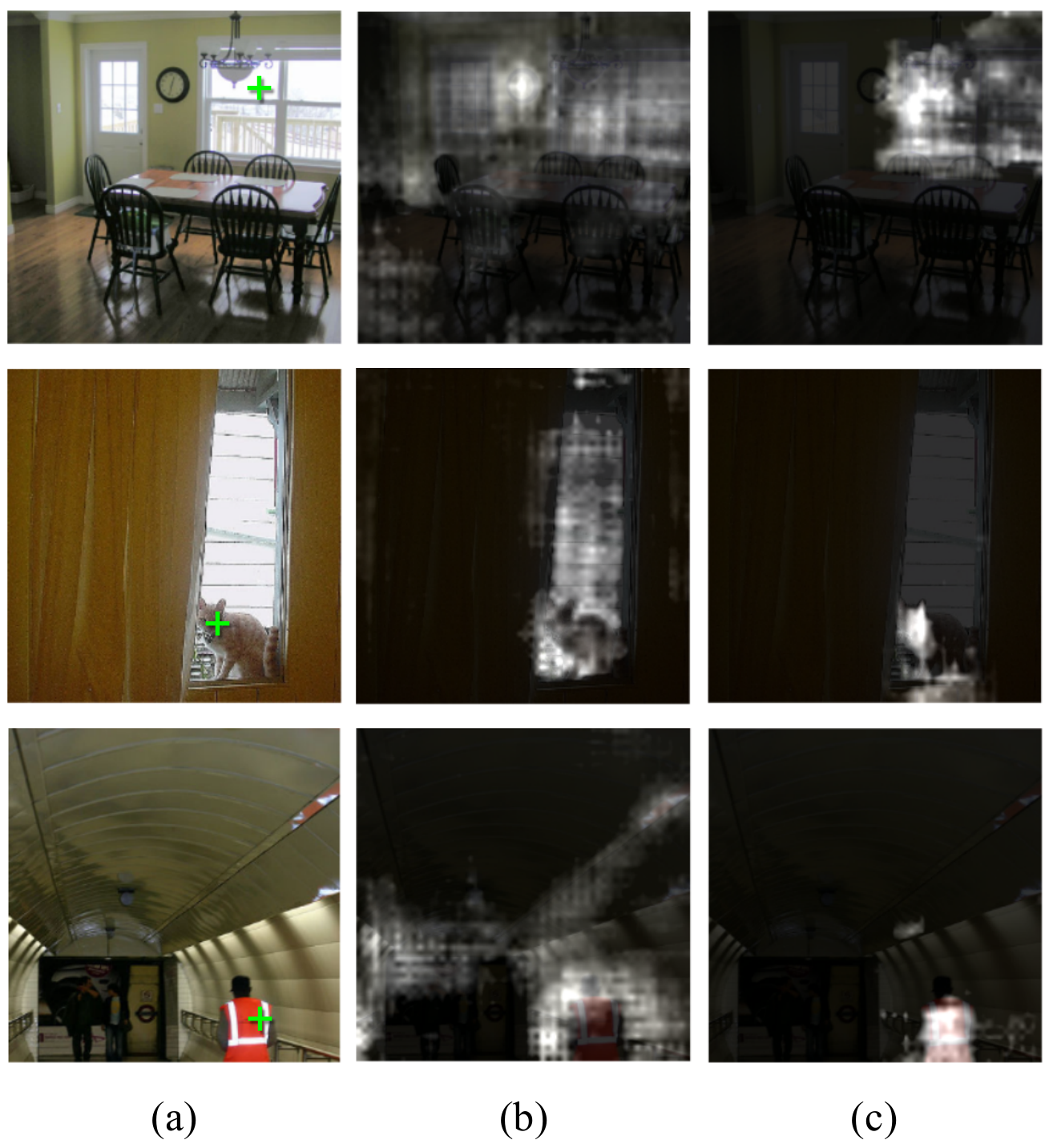}
  \caption{Visualization of attention maps. (a) Original images. (b) Attention maps for a certain position (green cross in (a)) computed with conventional self-attention. (c) Attention maps for a certain position (green cross in (a)) when self-attention is restricted in local patches. The attention maps of self-attention in local patches focus on surrounding areas that closely correlate with the specified position in (a).}
  \label{fig:patch_attn}
\end{figure}

Self-attention mechanism~\cite{wang2018non} is able to capture long-range dependencies between positions explicitly and has been applied to scene parsing~\cite{fu2019dual}\cite{zhang2019co} with a remarkable performance boost.
The key idea behind self-attention is that the response at a certain position is a weighted sum of features at all the positions. 
In this way, all the positions are related to each other and provide the network with a global receptive field. 
Long-range dependencies captured by self-attention can be combined with short-range ones captured by local convolution, leading to rich context information for dense labeling problems. 

However, long-range dependencies do not always work well for scene parsing tasks since a position is often less correlated with the positions far away from it, compared with those which are nearer. 
Moreover, information from distant positions may not be beneficial to building discriminative features. 
In Fig.~\ref{fig:patch_attn}(b), we visualize the attention map between a certain position and all the positions in the image computed in the self-attention process, where brighter color represents higher attention weight. 
The specified position is denoted as a green cross in Fig.~\ref{fig:patch_attn}(a). 
We find that in the conventional self-attention mechanism, the feature of this position would aggregate information from a wide area of the input image.
For example, in the second row of (b), the position on the cat receives information from distant ones of the image, like those on the window and curtain. 
However, there is no apparent correlation between the window, curtain, and cat. 
Thus the long-range dependencies captured from these positions are not useful for the model to classify a certain position on the cat.
When we restrict self-attention to a local patch of the image, the visualized attention map is more concentrated on the surrounding area of the specified position. 
For example, in the second row of Fig.~\ref{fig:patch_attn} (c), self-attention is restricted to the bottom-right $1/4$ patch of the image. 
The attention map between the specified position and all the positions within the patch mainly focuses on the head and body parts of the cat, indicating that useful middle-range dependencies among positions are captured. 
Benefiting from the close correlations within the same category, context information aggregated from these dependencies can severe as more valid guidance for the model to classify the specified position. 
Based on this observation, we devise a novel Middle-Range (MR) branch which restricts the self-attention mechanism to local patches of the input feature in order to fill the gap between long-range and short-range dependencies for complete context extraction.

Furthermore, we analyze the attention map generated with conventional self-attention and find that each position contributes different attention weights to others for context aggregation.
For each position, the total value of attention weights that it contributes to others reveals its correlation with other positions as well as its importance to the global context. 
A larger value implies that the position has stronger correlations with most of the other positions. 
Thus, the features of positions contributing higher attention weights to others encode the common patterns of the whole image, including main elements appearing in the scene, large-area continuous background, etc. 
These patterns contain useful global context information which is crucial to scene understanding.
By emphasizing features of the positions with larger contributions, long-range dependencies can be captured more accurately and adaptively to complicated scene elements, which can enhance the aggregation of global context by self-attention. 
We instantiate this idea with a Reweighed Long-Range (RLR) branch to modulate feature responses according to the attention weights contributions of each position.

With the newly proposed MR and RLR branches, we build an Omni-Range Dependencies Network (ORDNet) in which short-range, middle-range, and reweighed long-range dependencies collaborate seamlessly to achieve adaptability to diverse spatial region contents and relationships in natural scene images. The ORDNet is general and can be applied to any FCN backbone for learning more discriminative feature representations.

Our main contributions are summarized as follows: 
\begin{itemize}

\item We devise a Middle-Range (MR) branch which explicitly captures middle-range dependencies within local patches of scene image, filling the gap between long-range and short-range dependencies. 
\item We also propose a Reweighed Long-Range (RLR) branch to emphasize features of the positions which encode common patterns, so that more accurate and adaptive long-range dependencies could be captured.
\item With the above two branches, we develop a novel Omni-Range Dependencies Network (ORDNet) which effectively integrates short-range, middle-range and reweighed long-range dependencies to extract comprehensive context information for accurate scene parsing. Our ORDNet outperforms previous state-of-the-art methods on three popular scene parsing benchmarks, including PASCAL-Context~\cite{mottaghi2014role}, COCO Stuff~\cite{caesar2018coco} and ADE20K~\cite{zhou2017scene} datasets, which well demonstrates its effectiveness.

\end{itemize}

\begin{figure*}[t]
  \centering
  \includegraphics[width=1.0\linewidth]{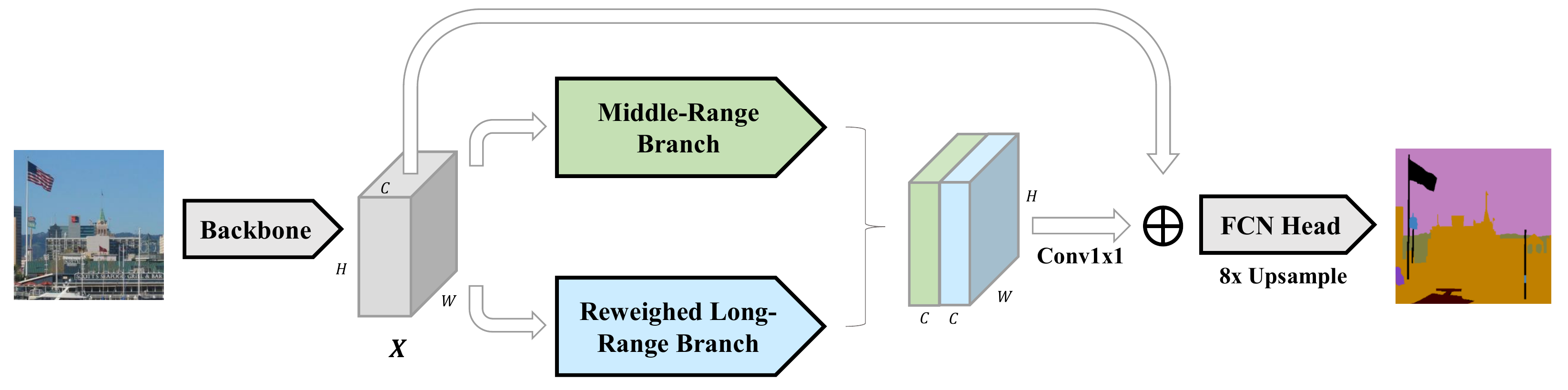}
  \caption{The pipeline of proposed Omni-Range Dependencies Network (ORDNet).  Given an input image, the extracted feature $X$ of a CNN backbone is fed into a Middle-Range (MR) branch and a Reweighed Long-Range (RLR) branch to capture the middle-range and reweighed long-range dependencies respectively. 
  The outputs of these two branches are then concatenated along the channel dimension and fused by a $1 \times 1$ convolution. An identity skip connection of $X$ is added to ease optimization. 
  The fused feature is fed into an FCN Head to predict the logit map and then upsampled $8$ times to obtain the final parsing mask.
 }
  \label{fig:pipeline}
\end{figure*}

\section{Related work}
\label{sec:related_work}

\subsection{Semantic Segmentation}
The goal of semantic segmentation is to assign category labels to the pixels of foreground objects and stuff in the scene, rather than segmenting the entire scene as scene parsing does. By expanding the set of pre-defined categories on which the model to segment, semantic segmentation serves as a basic technology of scene parsing. 
In recent years, remarkable progress has been achieved based on Fully Convolutional Networks~\cite{Long_2015_CVPR} (FCNs). FCN replaces the fully-connected layers of the image classification network (e.g., VGG16~\cite{simonyan2014very}) with convolution layers and introduces transposed convolution and skip layers to predict pixel-level labels. 
The many pooling layers in FCN increase the receptive field of convolution filters, and meanwhile reduce the resolution of feature maps, leading to inaccurate semantic masks. 
In order to maintain the resolution of feature maps while enjoying the increased receptive field of convolution filters, DeepLab~\cite{chen2014semantic} integrates atrous convolution into CNN, which boosts segmentation performance largely and becomes the de-facto component of latter segmentation methods.

Many deep models~\cite{kuanar2019adaptive}\cite{kuanar2018cognitive}\cite{kuanar2019low} have proposed various approaches to aggregate local spatial context information to refine the feature representations and achieved great performances.
Later works further propose to aggregate the important multi-scale spatial context information based on the last feature map of an FCN backbone.
For example, DeepLab v2~\cite{chen2017deeplab} employs
parallel atrous convolutions with different atrous rates called ASPP to capture context information of multiple receptive fields. 
DeepLab v3~\cite{chen2017rethinking} further integrates image-level features into ASPP to get a global receptive field. 
PSPNet~\cite{zhao2017pyramid} performs pooling operations at multiple grid scales in order to aggregate multi-scale contextual information. 
In addition to extracting context information from feature maps, 
EncNet~\cite{zhang2018context} uses a Context Encoding Module which exploits semantic category prior information of the scenes to provide global contexts. 
Recent DANet~\cite{fu2019dual} and CFNet~\cite{zhang2019co} exploit the self-attention mechanism to effectively capture long-range dependencies, which outperform previous multi-scale context aggregation methods in semantic segmentation and scene parsing. 
InterlacedSSA~\cite{Huang2019InterlacedSS} proposes a factorized self-attention approach to approximately capture long-range dependencies with low computational costs, which achieves comparable performance with DANet and CFNet. 
Different from the above methods, in this paper we further propose to capture middle-range dependencies and reweighed long-range dependencies to provide richer semantic information than vanilla self-attention. Our Omni-Range Dependencies Network (ORDNet) can fill the semantic gap between original long-range and short-range dependencies, and also capture more accurate long-range dependencies by feature reweighing, achieving more comprehensive scene understanding.

\subsection{Attention Mechanism}
Attention mechanism is first introduced in~\cite{bahdanau2014neural} for neural machine translation, and later widely applied to various tasks like machine translation~\cite{luong2015effective}, VQA~\cite{lu2016hierarchical}\cite{zhu2016visual7w}\cite{yang2016stacked} and image captioning~\cite{xu2015show}.
\cite{vaswani2017attention} is the first work to apply self-attention for capturing long-range dependencies within input sentences, achieving noticeably boosted performance in machine translation.
In~\cite{wang2018non}, self-attention mechanism is further extended to vision tasks and a non-local network is proposed to capture long-range dependencies.
For image tasks, self-attention methods compute the response at a position as a weighted sum of the features at all positions in the input feature maps, in which way the receptive field for the current position can go beyond local convolution kernels to cover the whole feature map.
Self-attention mechanism is widely adopted in vision tasks such as semantic segmentation~\cite{fu2019dual}~\cite{huang2018ccnet}, GANs~\cite{zhang2018self} and image de-raining~\cite{li2018non}.
Inspired by~\cite{wang2018non}, we propose a new self-attention architecture to capture omni-range dependencies of positions, where the Middle-Range (MR) branch restricts self-attention to patches to model middle-range dependencies and the Reweighed Long-Range (RLR) branch further emphasizes features of positions which encode common patterns of the image to obtain more accurate long-range dependencies. 
Compared with previous works, our method conforms better to practical spatial relations between semantic regions and achieves higher performance on several benchmarks.

In addition to self-attention, researchers also explore other attention methods to refine feature maps by adjusting their scales. 
SENet~\cite{hu2018squeeze} utilizes a squeeze-and-excitation process to recalibrate feature channels with signals pooled from the entire feature maps. 
The first squeeze operator conducts global average pooling to generates a channel descriptor as the global information embedding.
The second excitation operator maps the channel descriptor to a set of channel-specific weights with two successive fully connection layers. 
Finally, the channel-specific weights are multiplied with original features to rescale the channel responses.
CBAM~\cite{woo2018cbam} and BAM~\cite{park2018bam} apply SE operation to both channel and spatial dimensions. 
Our proposed reweighed long-range branch serves as a spatial recalibration module to some extent. 
Compared with the spatial branch in CBAM, which is based on the single response of the current position, our RLR branch reweighs the feature responses according to correlations among all positions of the entire feature map so that positions encoding common patterns and main elements of the scene can be emphasized to form a more discriminative feature representation.

\begin{figure*}[t]
  \centering
  \includegraphics[width=0.75\linewidth]{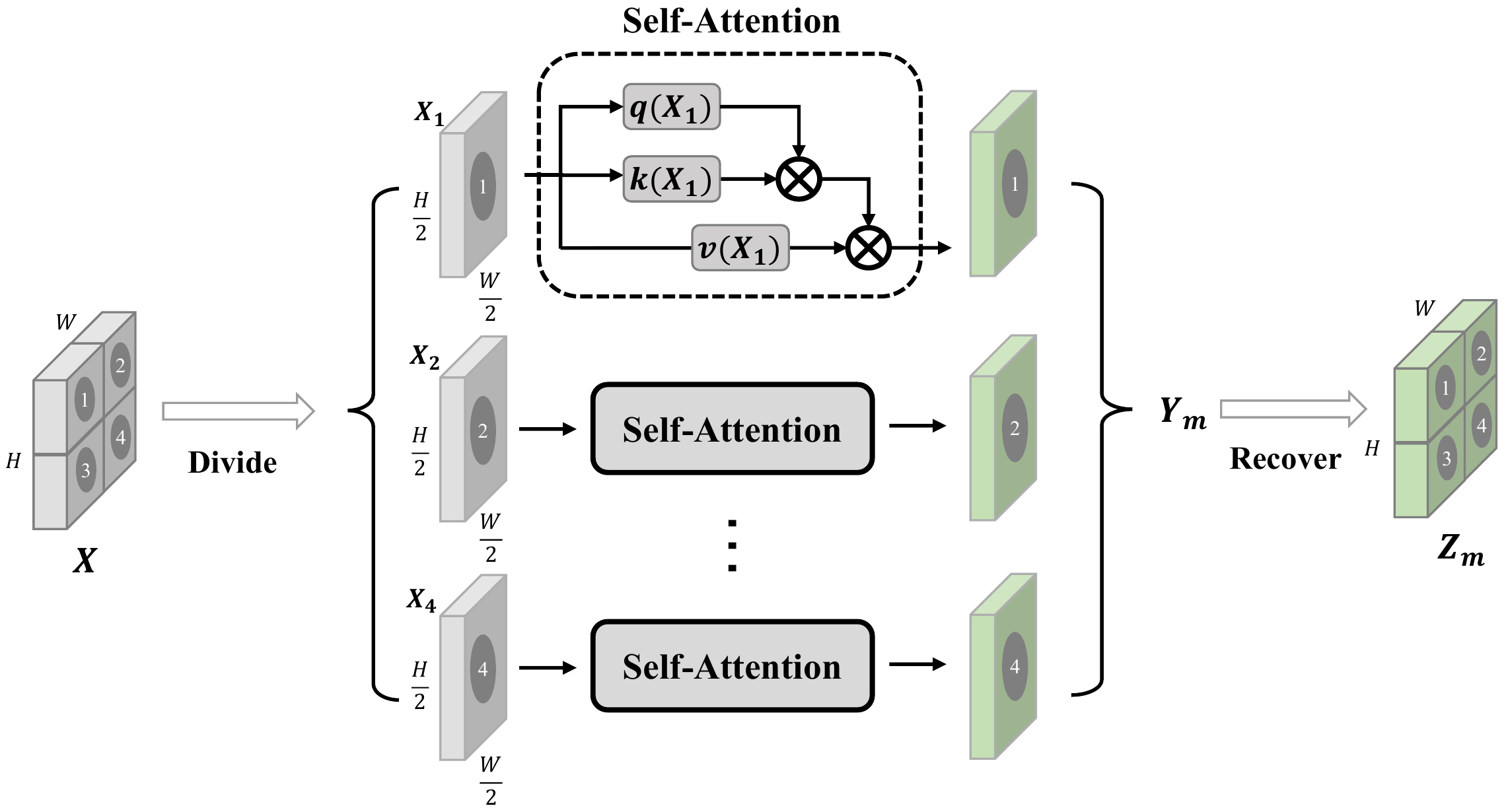}
  \caption{ 
  The pipeline of proposed Middle-Range (MR) branch. This branch contains 3 steps. First, The input feature $X \in \mathbb{R} ^ {H \times W \times C}$ is divided into $2 \times 2$ patches i.e. [$X_1$, $X_2$, $X_3$, $X_4$], which are ordered by rows. 
  Second, each patch is enhanced by a self-attention module separately to get $Y_{m} \in \mathbb{R}^{4 \times \frac{H}{2} \times \frac{W}{2} \times C}$ as the intermediate output. The operation of self-attention is as same as that described in \ref{sec:self-attention revisiting}. 
  Third, $Y_{m}$ is recovered to the same size of input feature to obtain $Z_{m}$ as final output.
  }
  \label{fig:mr_branch}
\end{figure*}

\section{Method}
\label{sec:method}
\subsection{Revisiting Self-Attention}
\label{sec:self-attention revisiting}
Self-attention mechanism computes the response at a position as a weighted sum of the features at all positions in the input feature maps. 
In this way, each position of the input feature can interact with others regardless of their spatial distance. 
The network can then effectively capture long-range dependencies among all the spatial positions. 
The overall workflow of self-attention is illustrated in the top area of Fig.~\ref{fig:mr_branch}. 
Given an input feature $X=[x^1; x^2; ...; x^{HW}]\in\mathbb{R}^{HW \times C}$ and an output feature $Y=[y^1; y^2; ...; y^{HW}]\in\mathbb{R}^{HW \times C}$, which are both reshaped to the matrix form, self-attention mechanism computes the output as
\begin{equation}
  \label{eq:self-attention formulation}
  y^i = \frac{1}{N(X)}\sum_{j=1}^{HW}attn^{ij}v(x^j),
\end{equation}
where $i$ is the index of a position of the output feature $Y$ and $j$ is the 
index enumerating all the positions of the input feature $X$. $H, W$ and $C$ are the height, width, and channel dimensions of $X$. 
$N(X)$ serves as a normalization factor which is set as $HW$.
$v(\cdot)$ is the value transform function implemented as a $1 \times 1$ convolution, $v(x^j) \in \mathbb{R}^{C_v}$ is the transformed feature. 
$attn^{ij}$ indicates the attention weight contributed by position $j$ to position $i$, which is defined as 
\begin{equation}
  \label{eq:relationship formulation}
  {attn^{ij}=q(x^i)^\top k(x^j)},
\end{equation}
where $q(\cdot)$ and $k(\cdot)$ are the query and key transform functions, $q(x^i) \in \mathbb{R}^{C_q}$ and $k(x^j) \in \mathbb{R}^{C_k}$ are transformed features respectively. 
In this work, we implement both $q(\cdot)$ and $k(\cdot)$ as $1 \times 1$ convolutions.
Eq.~(\ref{eq:self-attention formulation}) computes the response at query position $i$ as a weighted sum of the features of all positions.

Conventional self-attention, as described above, can effectively capture long-range dependencies, which well complements convolution layers that capture short-range dependencies within a local region. 
However, exploiting context information from too distant positions is sometimes problematic since a position is often less correlated with those far away from it. 
A position tends to have stronger correlations with those are moderately near from it, where middle-range dependencies are contained within this context. 
Therefore, it is necessary to fill the gap between short-range and long-range dependencies and capture various correlations among entities in the scene. 
Also, during the process of self-attention operation, the features of the positions which contribute significant attention weights to others usually encode common patterns contained in the scene. These patterns are beneficial to the comprehensive understanding of
sophisticated scenes and deserve appropriate emphasis for better capturing long-range dependencies.
Based on the two observations, we propose a novel Omni-Range Dependencies Network (ORDNet) which consists of a Middle-Range (MR) branch and a Reweighed Long-Range (RLR) branch to mine middle-range dependencies and refine long-range dependencies respectively for better scene understanding (\textit{see Fig.~\ref{fig:pipeline}}). 
We will elaborate on the MR branch in~\ref{sec:mr} and the RLR branch in~\ref{sec:rlr}. 
The overall architecture of ORDNet will be described in detail in~\ref{sec:ordnet}.

\subsection{Middle-Range Branch}
\label{sec:mr}
\begin{figure}[!htbp]
  \centering 
  \includegraphics[width=1.00\linewidth]{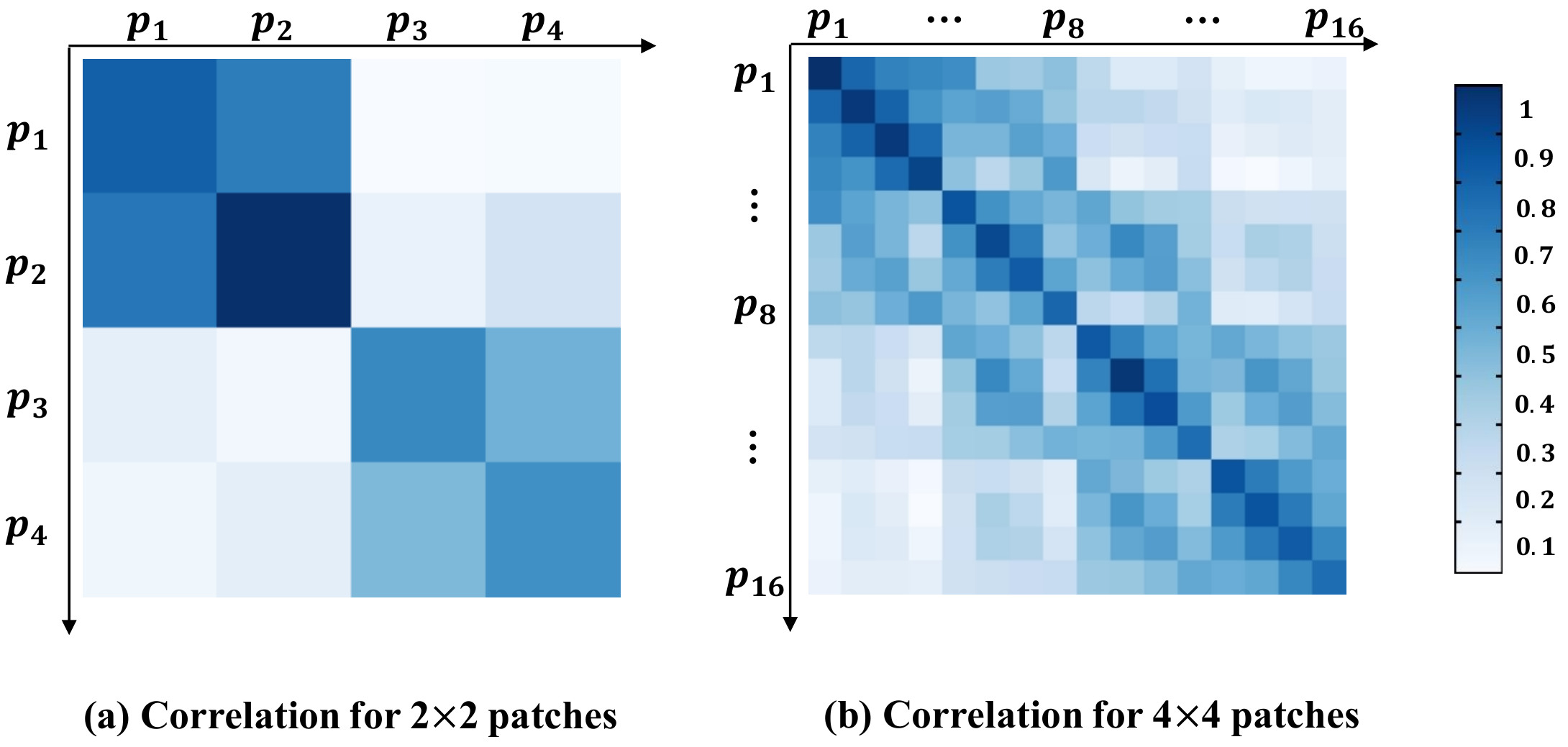}
  \caption{Visualization of intra-patch and inter-patch correlations. 
  (a) Correlation values computed on $2 \times 2$ patches.
  (b) Correlation values computed on $4 \times 4$ patches. The patches are ordered along 
  rows. Darker color denotes a larger correlation value. 
  The values of top left and bottom right elements as well as their surroundings are much larger than those of the rest regions.
  This means the pixels within the same or closer patches tend to have the same label. Furthermore, intra-patch correlation is usually stronger than inter-patch correlation.}
  \label{fig:patch_vis}
\end{figure}

Conventional self-attention exploits all positions of a feature map to update the feature of each query position. 
By analyzing correlation patterns among ground-truth mask patches in Fig.~\ref{fig:patch_vis}, 
we observe that a query position tends to have stronger correlations with the positions near to it compared with those which are distant. 
To demonstrate this, we randomly select $1,000$ images from PASCAL-Context~\cite{mottaghi2014role} dataset and divide the ground-truth masks into $2 \times 2$ and $4 \times 4$ patches along the height and width dimensions. 
We further calculate correlations inside and between patches by taking the average of corresponding similarity values:
\begin{equation}
  \label{eq:correlation function}
  Corr(p^m, p^n) = \sum_{i \in \Omega_{p^m}, j \in \Omega_{p^n}}sim(l^i, l^j)
  / (\vert\Omega_{p^m}\vert \vert\Omega_{p^n}\vert),
\end{equation}
where $sim(\cdot,\cdot)$ computes the similarity between two positions, which is either 1 or 0. We define the similarity between a pair of positions as whether their semantic labels are the same:
\begin{equation}
  \label{eq:similarity matrix}
  sim(l^i, l^j)=
  \begin{cases}
    1& l^i = l^j \\
    0& l^i \neq l^j.
  \end{cases}
\end{equation}
Here $l^i$ and $l^j$ denote the ground-truth category labels of position $i$ and $j$;
$p^n$ and $p^m$ represent the $n$-th and $m$-th patch, and $\Omega_{p^n}$ and $\Omega_{p^m}$ denote the set of all the positions belonging to them respectively; $\vert \cdot \vert$ means cardinality of their sets.

The visualized correlation matrices are shown in Fig.~\ref{fig:patch_vis} (a) and (b). 
We can observe that the values of the top left and bottom right elements, as well as their surroundings, are much larger (darker color) than those of the rest regions, which indicates intra-patch correlations are stronger than inter-patch ones. 
Moreover, pixels within the same or near patches tend to share the same label.
Visualization of the attention map in Fig.~\ref{fig:patch_attn} also demonstrates that conducting self-attention in local patches can capture middle-range dependencies among nearby similar positions instead of original long-range dependencies among all the positions.

Middle-range dependencies captured from local patches are able to provide more relevant context information than long-range dependencies, considering the higher intra-patch correlations than inter-patch ones.
We then develop a Middle-Range (MR) branch to capture such more informative middle-range dependencies to complement with long-range dependencies.
As illustrated in Fig.~\ref{fig:mr_branch}, our MR branch explicitly divides the input feature maps into $2\times 2$ patches and conducts self-attention operation within each patch separately. 
The output $Y_{m}$ is then recovered to the original spatial dimensions.
We narrow the self-attention range from the entire feature map to patch level so that the local nature of the feature can be exploited. Benefited from the complementarity among short-range, middle-range and long-range dependencies, 
the network can adapt to diverse spatial relationships between different scene elements.
We use $2 \times 2$ patches for all the experiments.  We also have tried to divide features into $4 \times 4$ patches but found diminishing return, possibly due to limited receptive field. 
Experimental results are shown in Tab.~\ref{tab:4x4patch}. 

\subsection{Reweighed Long-Range Branch}
\label{sec:rlr}
During the self-attention process, the multiplication between query and key features will generate an attention map where each element indicates the attention weight between each pair of positions. 
We observe that some positions contribute larger attention weights to other positions, which implies that there are stronger correlations between these positions and other ones.
Features of these positions usually encode common patterns like main elements appearing in the scene and large-area continuous background. 
These positions may be crucial to the global context during the self-attention process. By emphasizing features of these essential positions, the long-range dependencies modeled by self-attention is able to be more accurate. 
Therefore, we propose a Reweighed Long-Range (RLR) branch to selectively enhance features of these positions which contribute large attention weights to others. 

\begin{figure}[htbp]
  \centering 
  \includegraphics[width=1.00\linewidth]{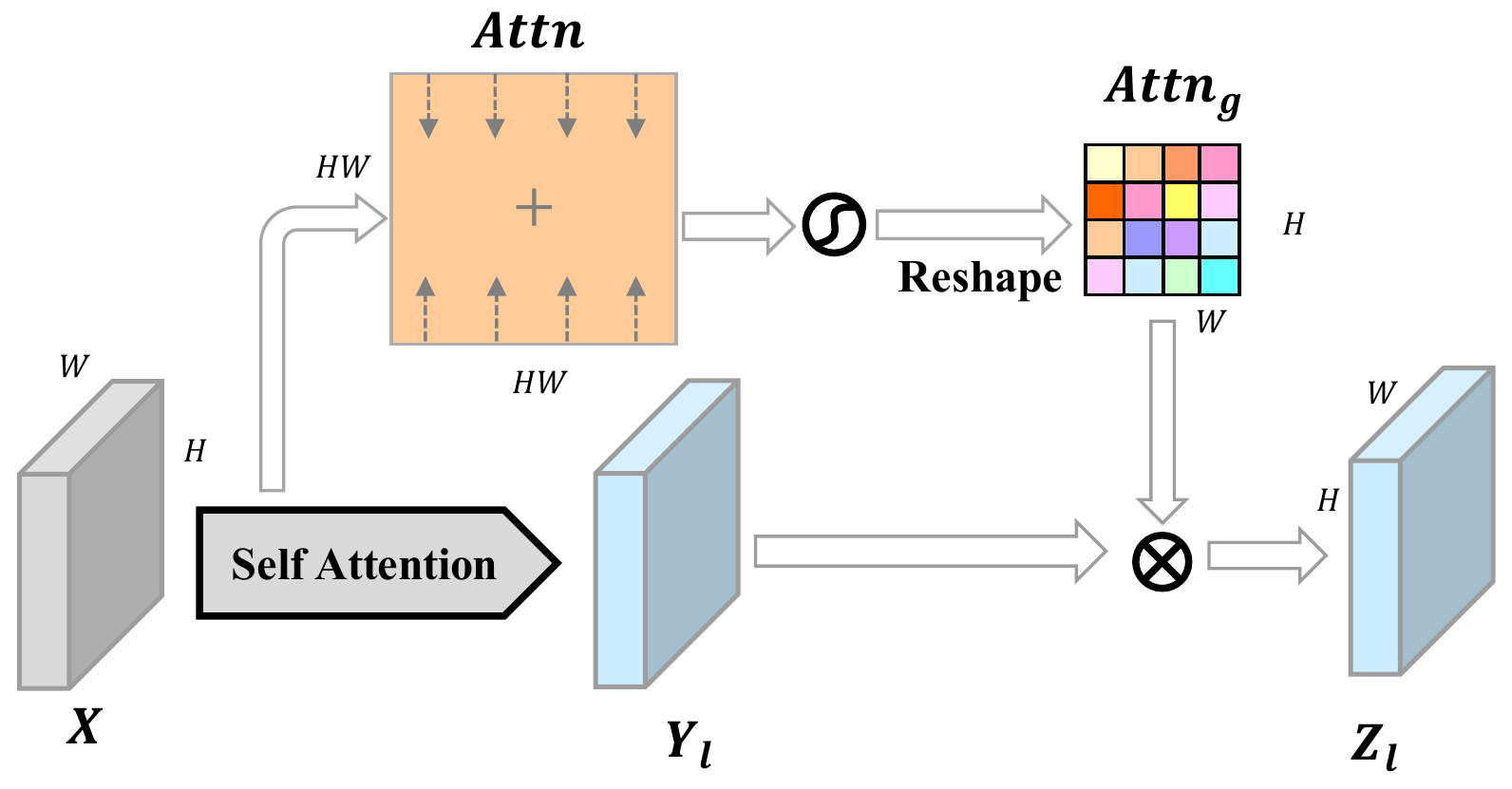}
  \caption{Our proposed Reweighed Long-Range (RLR) branch. 
  A self-attention module takes backbone feature as input $X$ and outputs attended feature $Y_{l}$ along with the attention map $Attn \in \mathbb{R}^{HW \times HW}$. 
  Then $Attn$ is summed up along each column and fed into a sigmoid function to get the attention contribution vector, which is then reshaped to $H \times W$ to obtain the global attention weight contribution map $Attn_g \in \mathbb{R}^{H \times W}$.  
  The output feature $Z_{l}$ is attained by multiplying $Attn_g$ with $Y_{l}$ elementwisely. 
  Note that the difference between $Attn$ and $Attn_g$ is that each element of $Attn$ denotes the attention weights between a pair of positions in $X$, whereas each element of $Attn_g$ denotes the summation of contribution of current position to all the positions.}
  \label{fig:rlr_branch}
\end{figure}

As illustrated in Fig~\ref{fig:rlr_branch}, given an input feature
$X \in \mathbb{R}^{H \times W \times C}$, we first feed it to a self-attention
module to output an attended feature $Y_{l} \in \mathbb{R}^{H \times W \times C}$ via 
Equation~(\ref{eq:self-attention formulation}) and the attention map $Attn \in \mathbb{R}^{HW \times HW}$
via Equation~(\ref{eq:relationship formulation}). $attn^{ji} \in Attn$ can be viewed as an
attention weight contributed by position $i$ to position $j$. We compute the global attention weight 
contribution at position $i$ by summing up all the attention weights it contributes to other positions and 
further employ a simple gated mechanism as below:

\begin{equation}
  \label{eq:spatial attention}
  attn^{i}_{g} = \sigma(\sum_{j=1}^{HW}attn^{ji}).
\end{equation}

Here $\sigma(\cdot)$ is the sigmoid function and is applied to normalize the scale of $Attn_g$. The spatial dimensions of the global attention weight contribution 
map $Attn_g=[attn^1_g; attn^2_g; ...; attn^{HW}_g]$ are $H\times W$ after reshaping operation. 
Through Equation~(\ref{eq:spatial attention}) we obtain $attn^{i}_g$, i.e., the normalized attention weight contribution 
of position $i$ to all the positions, which measures the effect of position $i$ on the global context of self-attended features. 
We finally multiply $Y_{l} \in \mathbb{R}^{H \times W \times C}$ with $Attn_g \in \mathbb{R}^{H \times W}$
elementwisely using the broadcasting rule to get $Z_{l} \in \mathbb{R}^{H \times W \times C}$ as the 
output of this branch:

\begin{equation}
  \label{eq:selective attention apply}
  Z_{l} = Y_{l} * Attn_g.
\end{equation}

By multiplied with the global attention weight contribution $Attn_g$, the feature at each position is reweighed according to its contribution to other positions. Features of positions which encode common patterns of the scene could be emphasized for better representation. 
Experimental results in Table~\ref{tab:Ablation} shows that our RLR branch improves the performance without introducing extra parameters. 

\subsection{Omni-Range Dependencies Network}
\label{sec:ordnet}
We here explain our proposed Omni-Range Dependencies Network (ORDNet) in detail. 
The architecture of our ORDNet is illustrated in Fig.~\ref{fig:pipeline}. 
We use ResNet101~\cite{he2016deep} pretrained on ImageNet~\cite{deng2009imagenet} as the backbone network to extract visual features. 
To enlarge the receptive field as well as maintain feature resolution, we replace the strided convolutions in the last two stages of ResNet with atrous convolutions, with stride set as $1$ and dilation rates set as $2$ and $4$ respectively. 
We also follow~\cite{zhang2018context} to replace the first $7 \times 7$ convolution of ResNet with $3$ consecutive $3 \times 3$ convolutions. 
The resolution of the output feature from backbone network $X \in \mathbb{R}^ {H \times W \times C}$ is $1/8$ of the original image. 
$X$ is then fed into our proposed two branches to capture middle-range and reweighed long-range dependencies in visual feature. 
After getting the output features $Z_{m}$ and $Z_{l}$ from these two branches, we concatenate them along the channel dimension. 
Then a $1\times 1$ convolution is applied on the concatenated feature to reduce its channel dimensions to the same number of the input feature. 
We also add a shortcut connection from input feature $X$ to the fused output of two branches to ease optimization. 
The fused output feature is then passed into an FCN head for final mask prediction, and we further upsample the prediction result by $8$ times to match the original resolution. 
In practice, our proposed two branches can be easily plugged into a segmentation network due to the residual nature and further enhance feature representations.

The concurrent work InterlacedSSA~\cite{Huang2019InterlacedSS} proposes a factorized self-attention method similar to our MR branch for semantic segmentation.
The motivation of InterlacedSSA is to decrease the computation/memory cost of self-attention mechanism by factorizing it into two consecutive self-attention processes occurred in patches. However, the motivation of our MR branch is to demonstrate the effectiveness of capturing middle-range dependencies by restricting self-attention in feature patches. Moreover, the factorized self-attention in InterlacedSSA still aims to capture long-range dependencies among positions by stepwise information propagation. However, the factorized self-attention in our MR branch aims to explicitly capture middle-range dependencies among positions to fill the semantic gap between long-range and short-range dependencies for more comprehensive scene understanding. It serves as an additional information source to complement with reweighed self-attention and normal convolutions. 
In summary, our ORDNet can make self-attention more comprehensive in aggregating information, while InterlacedSSA can reduce the computational budget of self-attention. 
InterlacedSSA makes a step forward over our middle-range branch and the contributions of us and InterlacedSSA are complementary.

\subsection{Loss Functions}
Our full loss function $\mathcal{L}_{Full}$ contains two parts namely standard cross-entropy loss $\mathcal{L}_{CE}$ and Lovasz-hinge loss~\cite{yu2015learning} $\mathcal{L}_{IoU}$, which are formulated as follows:

\begin{equation}
  \label{ce_loss}
  \mathcal{L}_{CE}(y^*, \tilde{y})=-\sum_{k = 1}^{K}{y_k^* \log{\tilde{y_k}}},
\end{equation}
\begin{equation}
  \begin{aligned}
  \label{iou_loss}
  & \mathcal{L}_{IoU}(y^*, \tilde{y}) = -Jaccard(y^*, \tilde{y}) \\
  & =-\sum_{k = 1}^{K}\frac{|(\arg\max (y^*) == k) \bigcap (\arg\max (\tilde{y}) == k)|}{|(\arg\max (y^*) == k) \bigcup (\arg\max (\tilde{y}) == k)|},
  \end{aligned}
\end{equation}
\begin{equation}
  \label{full_loss}
  \mathcal{L}_{Full}(y^*, \tilde{y})=\alpha_1\mathcal{L}_{CE}(y^*, \tilde{y}) + \alpha_2\mathcal{L}_{IoU}(y^*, \tilde{y}),
\end{equation}
where $y^*$ denotes ground-truth mask, $\tilde{y}$ denotes predicted logits, $\alpha_1$ and $\alpha_2$ are weights for different loss terms, $K$ denotes the number of semantic categories.\newline

\section{Experiments}
\subsection{Experimental Setup}
\textbf{Training: }
We conduct all the experiments using PyTorch~\cite{paszke2019pytorch} 
on three scene parsing benchmarks, 
including PASCAL-Context~\cite{mottaghi2014role}, COCO Stuff~\cite{caesar2018coco} and ADE20K~\cite{zhou2017scene}.
We also evaluate our model on PASCAL VOC 2012 dataset~\cite{Everingham10} for semantic segmentation task. 
We choose dilated FCN \cite{yu2015multi} as the baseline model and plug our MR and RLR branches between the ResNet backbone and FCN head. 
The output prediction is bilinearly upsampled by $8$ times to match the input size.
We initialize the backbone with an ImageNet~\cite{krizhevsky2012imagenet} pretrained model and other layers including MR branch, RLR branch and the FCN head are randomly initialized. 
We adopt the SGD optimizer with momentum set to $0.9$ and weight decay set to $0.0001$ to train the network.
We use the polynomial learning rate scheduling $lr=baselr*{(1-\frac{iter}{total\_iter})}^{power}$.
The base learning rate is set to $0.004$ for ADE20K dataset and $0.001$ for other datasets.
The channel dimension $C$ of the input feature $X$ for our MR branch and RLR branch is $2048$. 
Self-attention contains three linear layers to transform the input feature, namely query (q), key (k) and value (v). 
To reduce the memory cost, we set the output channel dimensions of query and key layers $C_q = C_k = 256$ and set $C_v = 512$ for all the self-attention modules we used. 
Our MR branch adopts $2 \times 2$ patches to capture middle-range dependencies. 
We conduct all the experiments on $4$ NVIDIA TITAN RTX GPU cards.
For data augmentation, We apply random flipping, 
random cropping and random resize between $0.5$ and $2$ for all datasets. 
When comparing with other methods, we adopt both standard cross-entropy loss and Lovasz-hinge loss~\cite{yu2015learning} as our full loss function to train the network. Loss weights $\alpha_1$ and $\alpha_2$ are both set as $1.0$. 
For the whole scene parsing dataset ADE20K, we follow~\cite{zhang2018context} and the
standard competition benchmark~\cite{zhou2017scene} to calculate mIoU by ignoring background pixels. 
When training our ORDNet on PASCAL-Context dataset, we also ignore the pixels of background category following~\cite{zhang2018context}\cite{fu2019dual}.

\textbf{Evaluation: }
As prior works~\cite{zhang2018context}\cite{fu2019dual}\cite{zhang2019co} show that employing 
multi-scale testing during evaluation is able to bring significant performance gain in semantic segmentation and 
scene parsing, we follow the best practice in~\cite{zhang2018context} to average the predictions of different scales as the 
final results. During evaluation, the input image is first resized according to a set of different scales 
$\{0.5, 0.75, 1.0, 1.25, 1.5, 1.75\}$, then cropped to a pre-defined image size for training. The cropped image is 
randomly flipped and fed into the segmentation network. The output logits are cropped and averaged across above 
scales as final prediction. For the consideration of fairness, we adopt multi-scale testing when comparing with 
state-of-the-art methods. Mean Intersection of Union (mIoU) and pixel accuracy (pixAcc) are adopted as evaluation
metrics.

\begin{figure*}[!htbp]
  \centering 
  \includegraphics[width=0.65\linewidth]{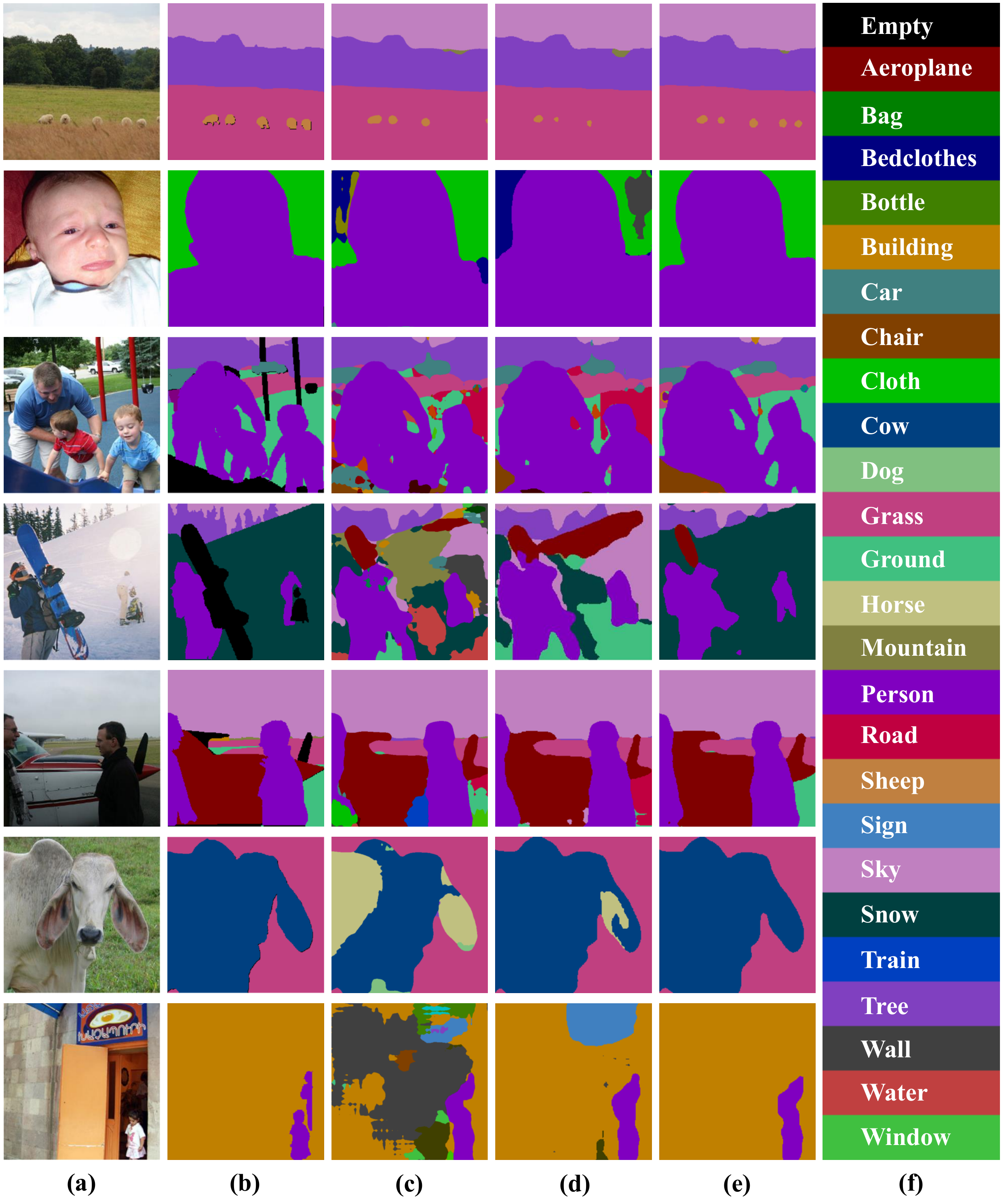}
  \caption{Qualitative comparison with baseline Dilated FCN and Basic SA on PASCAL-Context test set. (a) Original image. 
  (b) Ground-truth masks. (c) Results of Dilated FCN. (d) Results of Basic SA. (e) Results of our ORDNet. (f) Legend of semantic categories.}
  \label{fig:result}
\end{figure*}

\subsection{Ablation Studies}
We conduct both quantitative and qualitative ablation experiments on the test set of PASCAL-Context dataset~\cite{mottaghi2014role} 
to verify the effectiveness of our MR and RLR branches and their variants. We train our model for $50$ epochs with batch size of 
$16$. Following~\cite{zhang2019co}, we use the most common $59$ categories of PASCAL-Context without the background category for 
ablation studies.

\begin{table}[!htbp]
  \caption{Ablation studies on PASCAL-Context test set. mIoU is calculated on $59$ categories 
  w/o background.}
  \label{tab:Ablation}
  \centering
  \begin{tabular}{lccc}
    \toprule
    Method     & Backbone     & mIoU(\%)    & pixAcc(\%) \\
    \midrule
    Dilated FCN \cite{yu2015multi} & ResNet50  & 45.30   & 75.34     \\
    Basic SA \cite{wang2018non}     & ResNet50 & 49.45   & 78.61      \\
    Basic SA + MR     & ResNet50 & 50.26   & 78.95      \\
    Basic SA + RLR    & ResNet50 & 50.28   & 78.89      \\
    ORDNet    & ResNet50 & \textbf{50.67}   & \textbf{79.35}      \\
    ORDNet    & ResNet101 & \textbf{53.03}   & \textbf{80.24}      \\
    \bottomrule
  \end{tabular}
\end{table}

\textbf{MR and RLR branches: }
Experimental results of our proposed two branches are illustrated in Table~\ref{tab:Ablation}.
The dilated FCN baseline yields mIoU of $45.30\%$. After combining with basic self-attention 
(Basic SA)~\cite{wang2018non}, the mIoU can increase significantly by $4.15\%$, which forms a strong baseline for 
our method. 
Upon the basic self-attention mechanism (Basic SA), adding our Middle-Range branch (Basic SA + MR)
is able to achieve $0.81\%$ improvement in mIoU, which demonstrates that more comprehensive scene context information 
can be extracted by integrating middle-range dependencies into the segmentation network to complement with long-range 
dependencies captured by original self-attention. Adding our Reweighed Long-Range branch (Basic SA + RLR) can also bring $0.81\%$ 
mIoU gain over the Basic SA baseline, indicating that emphasizing features of positions which encode the common patterns of scenes is able to capture more accurate long-range dependencies to achieve better understanding of scenes. 
Furthermore, the RLR branch introduces no extra parameters over the strong Basic SA baseline while outperforming 
it by a large margin. When incorporating our proposed two branches together, our Omni-Range Dependencies Network (ORDNet) is 
capable of obtaining a further improvement with $1.22\%$ mIoU gain and $0.74\%$ pixAcc gain due to the 
effective complementarity between short-range (captured by convolutions), middle-range and reweighed long-range 
dependencies. After utilizing a deeper backbone network, our ORDNet can further achieve the performance boost to $3.58\%$ and 
$1.63\%$ gains in mIoU and pixAcc, respectively.

\textbf{Qualitative results: }
Qualitative comparison with baseline Basic SA~\cite{wang2018non} are shown 
in Fig~\ref{fig:result}. 
Our ORDNet obtains better parsing results in both global and local parts. 
For example, in the $4$-th row of (d) and (e), Basic SA produces muddled result with nonexistent categories, while our ORDNet is able to understand the entire scene correctly and generate coherent parsing map in the assistance of omni-range dependencies. 
Also in the $1$-st and $6$-th rows of (d) and (e), our ORDNet can fix local prediction errors in Basic SA, e.g., missing tiny sheeps and confusing ear, which indicates the superiority of integrating middle-range and reweighed long-range dependencies into the segmentation network. 
There are also some failure cases in Fig~\ref{fig:result}. 
For instance, our model fails to identify the ``background'' category (two pillars, snowboard) in the $3$-rd and $4$-th row. 
The reason is that we ignore the pixels of ``background'' category when training our ORDNet on PASCAL-Context dataset following~\cite{zhang2018context}\cite{fu2019dual}.  
Therefore, our ORDNet cannot identify the ``background'' category and predicts other labels for these pixels, which has no harm to the performance. 
We also observe other poor results like inaccurate boundaries in the last row of Fig~\ref{fig:result}. 
We suppose the reason is that our ORDNet cannot aggregate enough boundary details from low resolution feature maps. 
Exploiting low level features from the CNN backbone may alleviate this problem.

\textbf{Number of patches in MR branch: }
We also conduct analysis on $2\times 2$ patches and $4\times 4$ patches in our MR branch only. 
Experimental results 
are shown in Table~\ref{tab:4x4patch}. Given the same input size, conducting self-attention over $2\times 2$ patches reduce FLOPs from $94.50$G to $62.27$G and reduce mIoU from $49.45\%$ to $48.75\%$. 
Conducting self-attention over $4\times 4$ patches 
could reduce FLOPs from $94.50$G to $51.17$G given the same input size. 
However, it does not work as well as original self-attention (MR\_nopatch) or $2\times 2$ patches MR branch. Enlarging crop size 
to $640$ has limited improvement but results in larger FLOPs which is larger than $2\times 2$ patches (Second row). 
We suppose that dividing feature map into too many patches for self-attention will bring about fragmented receptive field, 
leading to inferior performance. Note that we only compare variants of MR branch without integrating with Basic SA, thus the mIoU and PixAcc of $2\times 2$ patches MR branch 
in Table~\ref{tab:4x4patch} are lower than those in Table~\ref{tab:Ablation}.

\begin{figure*}[t]
  \centering
  \includegraphics[width=0.75\linewidth]{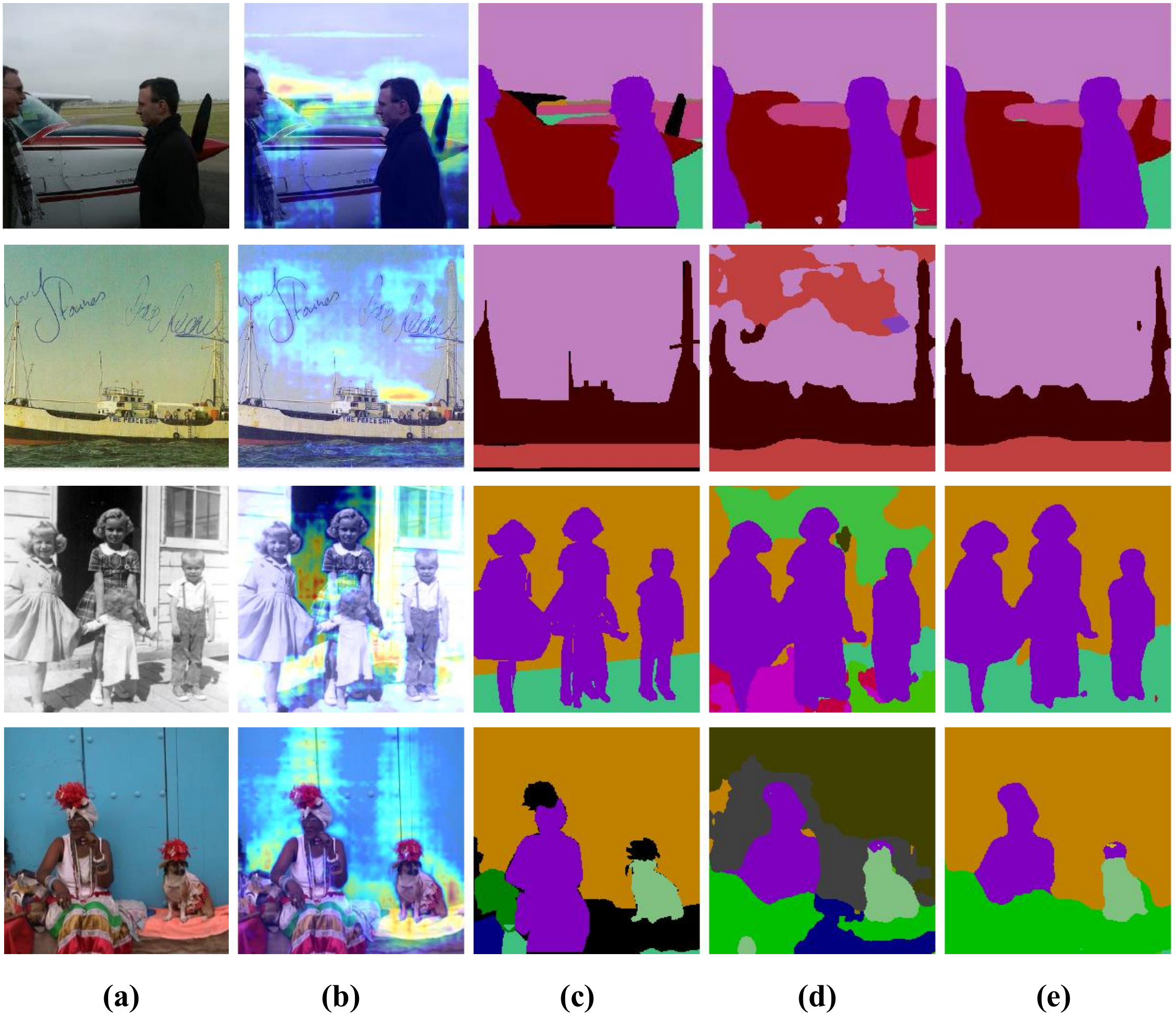}
 \caption{Visualization of attention maps and parsing results of self-attention 
 method and our RLR branch. (a) Original images. (b) Visualization of global attention weight contribution map $Attn_g$, which is calculated by summing up attention 
 weights that each position contributes to other positions. (c) Ground-truth label maps. (d) Parsing results of basic self-attention. 
 (e) Parsing results of RLR branch. We can observe from (d) that large-area continuous background usually contributes more attention weights to other positions, e.g. sky in the 2nd row, wall in the 4th row. 
 By emphasizing these regions, our RLR branch could correct the prediction errors made by basic-attention and assign proper labels to these regions. \textit{Best viewed in color.}}
  \label{fig:attention_out}
\end{figure*}

\begin{table}[!htbp]
  \caption{Results of different versions of MR branch on PASCAL-Context test set. 
  All the models are based on ResNet50 backbone.}
  \label{tab:4x4patch}
  \centering
  \begin{tabular}{lcccc}
    \toprule
    Method  & Crop Size & mIoU(\%) & PixAcc(\%) & GFLOPs \\
    \midrule
    MR\_nopatch & $480\times 480$ & \textbf{49.45} & \textbf{78.61} & 94.50 \\
    MR\_2x2patch & $480\times 480$ & 48.75  & 76.83 & 62.27 \\
    MR\_4x4patch & $480\times 480$ & 46.19  & 75.94 & 51.17 \\
    MR\_4x4patch & $640\times 640$ & 46.92 & 76.08 & 78.97 \\
    \bottomrule
  \end{tabular}
\end{table}

\begin{table}[!htbp]
  \caption{Results of different versions of RLR branch on PASCAL-Context test set.
  All the models are based on ResNet50 backbone.}
  \label{tab:rlr_versions}
  \centering
  \begin{tabular}{lccc}
    \toprule
    Attention Matrix  & Normalizing Method  & mIoU(\%) & PixAcc(\%) \\
    \midrule
    Attention-in & Softmax & 49.80 & 78.58  \\
    Attention-in & Sigmoid & 49.97  & 78.61 \\
    Attention-out & Softmax & 50.13 & \textbf{78.92}  \\
    Attention-out & Sigmoid & \textbf{50.28} & 78.89  \\
    \bottomrule
  \end{tabular}
\end{table}

\begin{table}[!htbp]
  \caption{Results of different fusing methods for outputs of MR branch and RLR branch. All the models are based on ResNet50 backbone.}
  \label{tab:fuse}
  \centering
  \begin{tabular}{lcc}
    \toprule
    Fusing method  & mIoU(\%) & PixAcc(\%) \\
    \midrule
    Element-wise Summation & 50.19 & 78.94  \\
    Concat + $1 \times 1$ Convolution & \textbf{50.67}  & \textbf{79.35} \\
    Attention to Scale~\cite{Chen2016Attention} & 49.73 & 78.81  \\
    Channel Selection~\cite{Li2019Selective} & 49.60 & 78.69  \\
    \bottomrule
  \end{tabular}
\end{table}

\textbf{Variants of RLR branch: }
We evaluate different versions of RLR branch. 
Experiment results are shown in Table~\ref{tab:rlr_versions}. All the models are based on ResNet50. Attention-in means 
that $Attn_g$ is obtained by summing up $Attn$ along each row so that each element of $Attn_g$ denotes the contribution 
of all positions to the current position. Attention-out means that  $Attn_g$ is obtained by summing up 
$Attn$ along each column so that each element of $Attn_g$ denotes the contribution of current position to all others. 
We also try different normalization methods for $Attn_g$, including Softmax and Sigmoid functions. 
Results in Table~\ref{tab:rlr_versions} show that calculating $Attn_g$ as attention-out attains better performance than 
attention-in and combining attention-out with sigmoid normalization achieves the best performance on PASCAL-Context dataset. 

\textbf{Fusing methods for MR and RLR branches: }
We compare different approaches to fuse the outputs of our proposed MR and RLR branches on the test set of PASCAL-Context.
Experiment results are presented in Table~\ref{tab:fuse}. All the models are based on ResNet50. Besides elementwise 
summation and concatenation followed by a $1 \times 1$ convolution, we also explore Attention to Scale~\cite{Chen2016Attention} 
and Channel Selection~\cite{Li2019Selective} for feature fusion. Attention to scale obtains a selection weight map for each branch and each 
position of the fused feature will receive a weighted summation of the features at the same positions from the two branches. 
Instead of spatial dimension, Channel Selection obtains the selection weight vector for each branch and conduct weighted summation 
along channel dimension. 
Experiment results indicate that simply concatenating the two features and fusing them with a $1 \times 1$ convolution 
achieves the best performance. A possible reason is that for features from our two branches, normal convolution fusing them 
along both spatial and channel dimensions, while Attention to Scale and Channel Selection perform fusion only along spatial dimension or 
channel dimension, respectively.

\textbf{Visualization of RLR branch: }
To further illustrate the proposed RLR branch, we visualize the attention maps and parsing results of our model 
with RLR branch only in Fig~\ref{fig:attention_out}. Column (b) visualizes of global attention weight contribution map $Attn_g$, i.e., summation of attention weights contributed by 
each position to all the positions. We can observe that areas with larger attention weight contribution (brighter color) usually represent common 
patterns of the scene, e.g., sky, grass and wall which serve as large-area continuous background. As shown in column (d) and (e), 
by emphasizing common pattern regions, our RLR branch is able to correct erroneous predictions made by basic self-attention and make the network understand the scene contents more comprehensively.

\begin{table}[!htbp]
  \caption{Comparison with state-of-the-art methods on PASCAL-Context test set. 
  mIoU is calculated on $60$ categories w/ background.}
  \label{tab:sota_pcontext}
  \centering
  \begin{tabular}{lccc}
    \toprule
    Method     & Backbone     & mIoU(\%) \\
    \midrule
    FCN-8s~\cite{Long_2015_CVPR} & -  & 37.8 \\
    CRF-RNN~\cite{zheng2015conditional} & - & 39.3 \\
    ParseNet~\cite{liu2015parsenet} & - & 40.4 \\
    BoxSup~\cite{dai2015boxsup} & - & 40.5 \\
    ConvPP-8~\cite{xie2015convolutional} & - & 41.0 \\
    HO\_CRF~\cite{arnab2016higher} & - & 41.3 \\
    PixelNet~\cite{bansal2016pixelnet} & - & 41.4 \\
    Piecewise~\cite{lin2016efficient} & - & 43.3 \\
    DAG-RNN + CRF~\cite{shuai2018scene} & - & 43.7 \\
    VeryDeep~\cite{wu2016bridging}  & -  & 44.5 \\
    DeepLab-v2~\cite{chen2017deeplab}    & ResNet101 + COCO    & 45.7 \\
    LabelBank~\cite{hu2017labelbank}  & ResNet101   & 45.8 \\
    RefineNet-101~\cite{lin2017refinenet}    & ResNet101  & 47.1 \\
    RefineNet-152~\cite{lin2017refinenet}    & ResNet152  & 47.3 \\
    PSPNet~\cite{zhao2017pyramid}     & ResNet101 & 47.8 \\
    Model A2, 2 conv~\cite{Wu2016Wider} & - & 48.1 \\
    MSCI~\cite{lin2018multi} & ResNet152 & 50.3 \\
    SGR~\cite{liang2018symbolic} & ResNet101 & 50.8 \\
    CLL~\cite{ding2018context}   & ResNet101  & 51.6 \\
    EncNet~\cite{zhang2018context}    & ResNet101 & 51.7 \\
    SGR+~\cite{liang2018symbolic} & ResNet101 + COCO Stuff & 52.5 \\
    DUpsampling~\cite{tian2019decoders} & Xception-71 & 52.5 \\
    DANet~\cite{fu2019dual} & ResNet101 & 52.6 \\
    SVCNet~\cite{ding2019semantic} & ResNet101 & 53.2 \\
    CFNet~\cite{zhang2019co}  & ResNet101 & 54.0 \\
    InterlacedSSA~\cite{Huang2019InterlacedSS} & ResNet101 & 54.1 \\
    \midrule
    ORDNet (ours)    & ResNet101 & \textbf{54.5} \\
    \bottomrule
  \end{tabular}
\end{table}

\subsection{Results on PASCAL-Context}
\label{sec:repc}
PASCAL-Context dataset~\cite{mottaghi2014role} contains $4,998$ images for training and $5,105$ images for testing. 
All images are densely annotated with $60$ categories in total including background.
We train our model for $80$ epochs with batch size of $16$ when comparing with state-of-the-art methods. 
The image crop size is set to $480$. 
We use the total $60$ categories including the most frequent $59$ categories and the background category to evaluate our method as prior works~\cite{lin2017refinenet}\cite{zhang2018context}\cite{zhang2019co} do.

Compared methods and results are shown in Table~\ref{tab:sota_pcontext}. 
Our ORDNet achieves $54.5\%$ mIoU which outperforms previous state-of-the-art methods. 
SGR+~\cite{liang2018symbolic} and DeepLab-v2~\cite{chen2017deeplab} utilize additional COCO Stuff~\cite{caesar2018coco} and COCO~\cite{lin2014microsoft} data to pretrain their models. 
RefineNet-152~\cite{lin2017refinenet} and MSCI~\cite{lin2018multi} adopt deeper backbone network to boost performance. 
Recent CFNet~\cite{zhang2019co} integrates the Context Encoding module from EncNet~\cite{zhang2018context} and obtains $54.0\%$ mIoU. 
InterlacedSSA~\cite{Huang2019InterlacedSS} further boosts the performance to $54.1\%$ mIoU via factorized self-attention similar to our MR branch. 
Our ORDNet achieves better performance than above methods without using extra pretraining data, deeper backbone network or other context aggregation modules.
It is demonstrated that capturing omni-range dependencies is more effective in providing richer semantic information for scene parsing.

\begin{table}[!htbp]
  \caption{Comparison with state-of-the-art methods on COCO Stuff test set.}
  \label{tab:sota_coco}
  \centering
  \setlength{\tabcolsep}{4mm}{
  \begin{tabular}{lccc}
    \toprule
    Method     & Backbone   & mIoU(\%) \\
    \midrule
    FCN~\cite{caesar2018coco} & -  & 22.7 \\
    FCN-8s~\cite{Long_2015_CVPR} & -  & 27.2 \\
    DAG-RNN + CRF~\cite{shuai2018scene}   & ResNet101  & 31.2 \\
    RefineNet~\cite{lin2017refinenet}    & ResNet152       & 33.6 \\
    LabelBank~\cite{hu2017labelbank}  & ResNet101   & 34.3 \\
    DeepLab-v2~\cite{chen2017deeplab}    & ResNet101       & 34.4 \\
    CLL~\cite{ding2018context}   & ResNet101  & 35.7 \\
    DSSPN~\cite{liang2018dynamic}   & ResNet101  & 36.2 \\
    SGR~\cite{liang2018symbolic}  & ResNet101   & 39.1   \\
    InterlacedSSA~\cite{Huang2019InterlacedSS} & ResNet101 & 39.2 \\
    SVCNet~\cite{ding2019semantic} & ResNet101 & 39.6 \\
    DANet~\cite{fu2019dual}    & ResNet101  & 39.7      \\
    \midrule
    ORDNet (ours)    & ResNet101    & \textbf{40.5}  \\
    \bottomrule
  \end{tabular}}
\end{table}

\begin{table}[!htbp]
  \caption{Comparison with state-of-the-art methods on ADE20K validation set.}
  \label{tab:sota_ade20k}
  \centering
  \begin{tabular}{lccc}
    \toprule
    Method         & Backbone  &  mIoU(\%) &  pixAcc(\%)   \\
    \midrule
    SegNet~\cite{Badrinarayanan2017SegNet}   & -    & 21.64     &  71.00   \\
    FCN~\cite{Long_2015_CVPR} &  -   & 29.39  & 71.32    \\
    DilatedNet \cite{yu2015multi} &  -  & 32.31  &  73.55   \\
    CascadeNet~\cite{zhou2017scene} &  -  & 34.90  &  74.52   \\
    DeepLabv2~\cite{chen2017deeplab} & ResNet101  & 38.97  &  79.01 \\
    RefineNet-101~\cite{lin2017refinenet}    & ResNet101   & 40.20  & -    \\
    RefineNet-152~\cite{lin2017refinenet}    & ResNet152   & 40.70  & -    \\
    DSSPN~\cite{liang2018dynamic}        & ResNet101   & 42.03  &  81.21    \\
    PSPNet-101~\cite{zhao2017pyramid}     & ResNet101      & 43.29  &  81.39    \\
    PSPNet-269~\cite{zhao2017pyramid}     & ResNet269      & 44.94  &  81.69    \\
    Model A2, 2c~\cite{Wu2016Wider}    & -  & 43.73   & 81.17  \\
    SGR~\cite{liang2018symbolic}    & ResNet101   & 44.32    & 81.43     \\
    EncNet~\cite{zhang2018context}    & ResNet101  & 44.65 & \textbf{81.69} \\
    CFNet~\cite{zhang2019co}          & ResNet101  & 44.89 & - \\
    InterlacedSSA~\cite{Huang2019InterlacedSS} & ResNet101 & 45.04 & - \\
    \midrule
    ORDNet (ours)    & ResNet101  & \textbf{45.39}  &  81.48   \\
    \bottomrule
  \end{tabular}
\end{table}

\begin{table*}[t]
  \caption{Results on ADE20K test set. Evaluation provided by the challenge organizers.}
  \label{tab:sota_ade20k_test}
  \centering
  \begin{tabular}{lcccc}
    \toprule
    Method         & Ensemble models & Train/Trainval & Backbone     & Final Score \\
    \midrule
    Dense Relation Network & - & -  & -  & 56.35 \\
    DRANet101\_SingleModel   & No & Trainval & ResNet101  & 56.72 \\
    Adelaide  & Yes & Trainval & - & 56.73 \\
    SenseCUSceneParsing   & No & Train &- &  55.38 \\
    SenseCUSceneParsing   & Yes & Trainval &- &  \textbf{57.21} \\
    \midrule
    ORDNet(Ours)  & No & Train & ResNet101 &  56.67 \\ 
    ORDNet(ours)  & No & Trainval &  ResNet101  & \textbf{56.86}   \\
    \bottomrule
  \end{tabular}
\end{table*}

\subsection{Results on COCO Stuff}
COCO Stuff dataset~\cite{caesar2018coco} contains
$10,000$ images from MSCOCO dataset~\cite{lin2014microsoft} with dense annotations of $80$ thing (e.g. book, clock) 
and $91$ stuff categories (e.g. flower). We use $9,000$ images for training and the rest for testing.
We adopt batch size of $16$ and train our model for $80$ epochs. The image crop size is set to $520$.
Mean IoU results calculated on all the $171$ categories are shown in Table~\ref{tab:sota_coco}. 
Among the compared methods, DAG-RNN~\cite{shuai2018scene} utilizes chain-RNNs to model rich spatial dependencies. 
CCL~\cite{ding2018context} adopts a gating mechanism in the decoder stage to improve inconspicuous objects and background stuff segmentation. 
SGR~\cite{liang2018symbolic} uses a knowledge graph to convert image features into symbolic nodes and conducts graph reasoning on them. 
SVCNet~\cite{ding2019semantic} generates a scale- and shape-variant semantic mask for each pixel to confine its contextual region for more adaptive context aggregation. 
DANet~\cite{fu2019dual} employs spatial and channel-wise self-attention to further improve performance.
Recent InterlacedSSA~\cite{Huang2019InterlacedSS} proposes a factorized approach similar to our MR branch to accelerate self-attention.
Our method outperforms these methods with a large margin and achieves a new state-of-the-art result of $40.5\%$ mIoU with no external knowledge used. 
This result indicates that capturing omni-range dependencies is more effective than 
merely modeling long-range dependencies in conventional self-attention.

\subsection{Results on ADE20K}
ADE20K~\cite{zhou2017scene} is a large scene parsing benchmark with $150$ categories including stuff and objects. 
It contains $20,210$ images for training and $2,000$ for validation. 
We train our model for $120$ epochs on the training set with batch size 
of $16$ and report mIoU and pixAcc results on the validation set. As the average image size of ADE20K dataset is larger than 
others, we adopt image crop size of $576$ on ADE20K. 
Comparison with previous state-of-the-art methods is shown in Table~\ref{tab:sota_ade20k}. 
PSPNet-269~\cite{zhao2017pyramid} uses a much deeper backbone network than other methods. EncNet~\cite{zhang2018context} and 
CFNet~\cite{zhang2019co} exploit prior information of semantic categories appearing in the scene to improve performance. 
InterlacedSSA~\cite{Huang2019InterlacedSS} introduces factorized self-attention similar to our MR branch and obtain $45.04\%$ mIoU.
Our method achieves $45.39\%$ mIoU which outperforms previous methods without using deeper backbone, category prior or external knowledge like~\cite{liang2018symbolic}. 
As mentioned in the Section $4.1$ of InterlacedSSA, the $0.2\%$ improvement of their method is not neglectable considering the improvements on ADE20K is very challenging.
Therefore, results of our method demonstrate capturing omni-range dependencies is also effective in challenging and complicated scenes.

To further demonstrate the effectiveness of our ORDNet, we evaluate our model on the test set.
The experiment results are shown in Table~\ref{tab:sota_ade20k_test}. Our model without finetuning on validation set achieves $56.67$ final score and surpasses most other methods. The final score denotes the average of PixAcc and mIoU.
For fair comparison, we further finetune our model on the train+val set of ADE20K for $20$ epochs, with the same training scheme except that the initial learning rate is set to $1e-4$.
Our ORDNet achieves a final score of $56.86$ on the test set with a single model and ranks at the $2$-nd place on the leaderboard of MIT Scene Parsing Benchmark. 
Our single model surpasses the $3$-rd place, Adelaide, which ensembles multiple models. 
The $1$-st place, SenseCUSceneParsing achieves $57.21$ final score by ensembling multiple models as well. 
Its single model trained only on the training set achieves $55.38$ while our ORDNet achieves $56.67$ with the same setting.

\subsection{Results on PASCAL VOC 2012}
We also evaluate the proposed ORDNet on PASCAL VOC 2012 dataset~\cite{Everingham10} with $21$ categories 
for semantic segmentation task. 
We adopt the augmented dataset~\cite{hariharan2011semantic} which contains $10,582$ images for training, 
$1,449$ images for validation and $1,456$ images for testing. 
We first train on the augmented train + val set for $60$ epochs with 
initial learning rate of $1e-3$. Then we finetune the model on the original PASCAL VOC training set for another 
$20$ epochs with learning rate of $1e-4$. We adopt ResNet101 as backbone network and the image crop size is set to $480$. 
Our ORDNet achieves $83.3\%$ mIoU on PASCAL VOC 2012 test set without using COCO pretraining or additional 
context aggregation modules. It is demonstrated that our method can also adapt to foreground object segmentation task by capturing 
omni-range dependencies.

\subsection{Analysis of Computational Overhead}
As shown in TABLE~\ref{tab:compute_cost}, all the models are run on a single NVIDIA TITAN XP GPU card to report computational overhead. 
InterlacedSSA is superior to other methods including ours in terms of speed and memory cost. 
It is a natural result since the main idea of InterlacedSSA is to decrease the computational budget of self-attention via feature factorization. 
While the main idea of our ORDNet is to demonstrate the effectiveness of capturing omni-range dependencies by our MR and RLR branches, which outperforms InterlacedSSA on all the datasets we adopted in this paper. 
Reducing computational complexity is not one of our claims. 
Comparing with Basic SA and DANet, our ORDNet actually has a moderate computational overhead since we reduce the channel dimensions of query, key and value layers in our MR and RLR branches to $256$, $256$ and $512$ respectively, but our ORDNet achieves higher performances. We will explore how to reduce its time and memory costs in the future work.

\begin{table}[!htbp]
  \caption{Efficiency comparison given input feature map of size $[2048 \times 128 \times 128]$ in inference stage.}
  \label{tab:compute_cost}
  \centering
  \begin{tabular}{lccc}
    \toprule
    Methods & Memory (MB) & GFLOPs & Time (ms) \\
    \midrule
    InterlacedSSA~\cite{Huang2019InterlacedSS} & \textbf{252} & \textbf{386} & \textbf{45} \\
    Basic SA~\cite{wang2018non} & 2168 & 619 & 77 \\
    Ours & 2192 & 624 & 83 \\
    DANet~\cite{fu2019dual} & 2339 & 1110 & 121 \\
    \bottomrule
  \end{tabular}
\end{table}

\section{Conclusion and future work}
In this paper, we address the scene parsing problem which requires the model to segment the entire scene instead of 
foreground objects. We propose a novel Omni-Range Dependencies Network (ORDNet) which restricts the scope of self-attention
to local patches to capture middle-range dependencies and meanwhile selectively emphasizes spatial regions 
contributing significant attention weights to others to model more accurate long-range dependencies. 
By integrating middle-range, reweighed long-range and short-range dependencies captured by local convolutions together, 
our ORDNet can aid models in adapting to various spatial scales and relationships in the complicated natural images, 
thus strengthening local and global feature representations. Extensive experiments on four scene parsing and segmentation 
benchmarks demonstrate its superior performance. Furthermore, our ORDNet can be applied to other visual tasks for capturing 
omni-range dependencies due to its generality and plug-and-play property. In the future, we hope to apply the ORDNet to other 
visual tasks and study how to further reduce its computing budget.

{
\bibliographystyle{IEEEtran}
\bibliography{ordnet_final}
}

\end{document}